\documentclass[10pt,twocolumn,letterpaper]{article}

\usepackage{wacv}              

\usepackage{graphicx}
\usepackage{amsmath}
\usepackage{amssymb}
\usepackage{booktabs}

\usepackage[linesnumbered,ruled,vlined]{algorithm2e}
\usepackage{multirow}
\usepackage{float}
\usepackage{graphicx}
\usepackage[accsupp]{axessibility}
\usepackage{stfloats}

\usepackage[pagebackref,breaklinks,colorlinks]{hyperref}

\usepackage[capitalize]{cleveref}
\crefname{section}{Sec.}{Secs.}
\Crefname{section}{Section}{Sections}
\Crefname{table}{Table}{Tables}
\crefname{table}{Tab.}{Tabs.}

\def\wacvPaperID{998} 
\def\confName{WACV}
\def\confYear{2024}

\begin{document}

\title{RecycleNet: Latent Feature Recycling Leads to Iterative Decision Refinement}

\author{Gregor Koehler\textsuperscript{1,2}, Tassilo Wald\textsuperscript{1,3}, Constantin Ulrich\textsuperscript{1.4}, David Zimmerer\textsuperscript{1,3},\\Paul F. Jaeger\textsuperscript{3,5}, Jörg K.H. Franke\textsuperscript{7}, Simon Kohl\textsuperscript{8}, Fabian Isensee\textsuperscript{1,3,6} and Klaus H. Maier-Hein\textsuperscript{1,3,4,9}\\
\\
\textsuperscript{1}German Cancer Research Center (DKFZ) Heidelberg, Division of Medical Image Computing, Germany\\
\textsuperscript{2}Helmholtz Information and Data Science School for Health, Karlsruhe/Heidelberg, Germany\\
\textsuperscript{3}Helmholtz Imaging, DKFZ\\
\textsuperscript{4}National Center for Tumor Diseases (NCT), NCT Heidelberg, \\a partnership between DKFZ and University Medical Center Heidelberg\\
\textsuperscript{5}Interactive Machine Learning Group, DKFZ \\
\textsuperscript{6}Applied Computer Vision Lab, DKFZ\\
\textsuperscript{7}Machine Learning Lab,  University  of  Freiburg, Freiburg, Germany\\
\textsuperscript{8}Latent Labs ({\hypersetup{urlcolor=black}\href{http://www.latentlabs.com}{latentlabs.com}}), London, UK\\
\textsuperscript{9}Pattern Analysis and Learning Group, Heidelberg University Hospital, Heidelberg, Germany\\
{\tt\small g.koehler@dkfz.de}
}
\maketitle

\begin{abstract}
   Despite the remarkable success of deep learning systems over the last decade, a key difference still remains between neural network and human decision-making: As humans, we can not only form a decision on the spot, but also ponder, revisiting an initial guess from different angles, distilling relevant information, arriving at a better decision. Here, we propose RecycleNet, a latent feature recycling method, instilling the pondering capability for neural networks to refine initial decisions over a number of recycling steps, where outputs are fed back into earlier network layers in an iterative fashion. This approach makes minimal assumptions about the neural network architecture and thus can be implemented in a wide variety of contexts. Using medical image segmentation as the evaluation environment, we show that latent feature recycling enables the network to iteratively refine initial predictions even beyond the iterations seen during training, converging towards an improved decision. We evaluate this across a variety of segmentation benchmarks and show consistent improvements even compared with top-performing segmentation methods. This allows trading increased computation time for improved performance, which can be beneficial, especially for safety-critical applications.
\end{abstract}

\section{Introduction}
\label{sec:intro}

Over the past decade, the field of computer vision has witnessed an unprecedented paradigm shift due to the advent and proliferation of deep learning algorithms. Neural networks have become the de facto standard for a variety of tasks, excelling in their ability to solve previously impossible tasks across many domains and modalities. One of the most intriguing distinctions between human cognition and neural networks, however, is the former's capacity for iterative decision-making - a skill that is still lacking from most recent artificial systems. Humans exhibit the innate ability to dynamically revisit and revise their initial decisions, evaluating their options from multiple perspectives, and continuously improving their decision based on new information. This process underlies an inherent characteristic of human decision-making – that of iterative refinement. It allows humans to evolve their understanding over time, improving the quality of decisions, especially in complex, non-deterministic scenarios. By stark contrast, conventional deep learning architectures have typically operated in a one-shot, feed-forward manner, lacking the property of iterative revision.\\
In this work, we seek to bridge this gap, to bring neural networks a step closer to the iterative decision-making process that characterizes human cognition. We introduce RecycleNet, a simple approach to deep learning that leverages the concept of latent feature recycling, enabling neural networks to refine their initial predictions over a series of iterative steps. The advantage of RecycleNet lies in its universal applicability - it makes minimal assumptions about the underlying architecture and is easily adaptable across a wide array of contexts. Medical image segmentation, with its critical implications for diagnosis and treatment in healthcare, serves as an excellent testing ground for our approach. The task's complexity and the inherent noise in medical imaging data pose formidable challenges that demand robust, reliable, and refined predictions. 

\begin{figure}[!ht]
\centering
\includegraphics[width=0.9\linewidth]{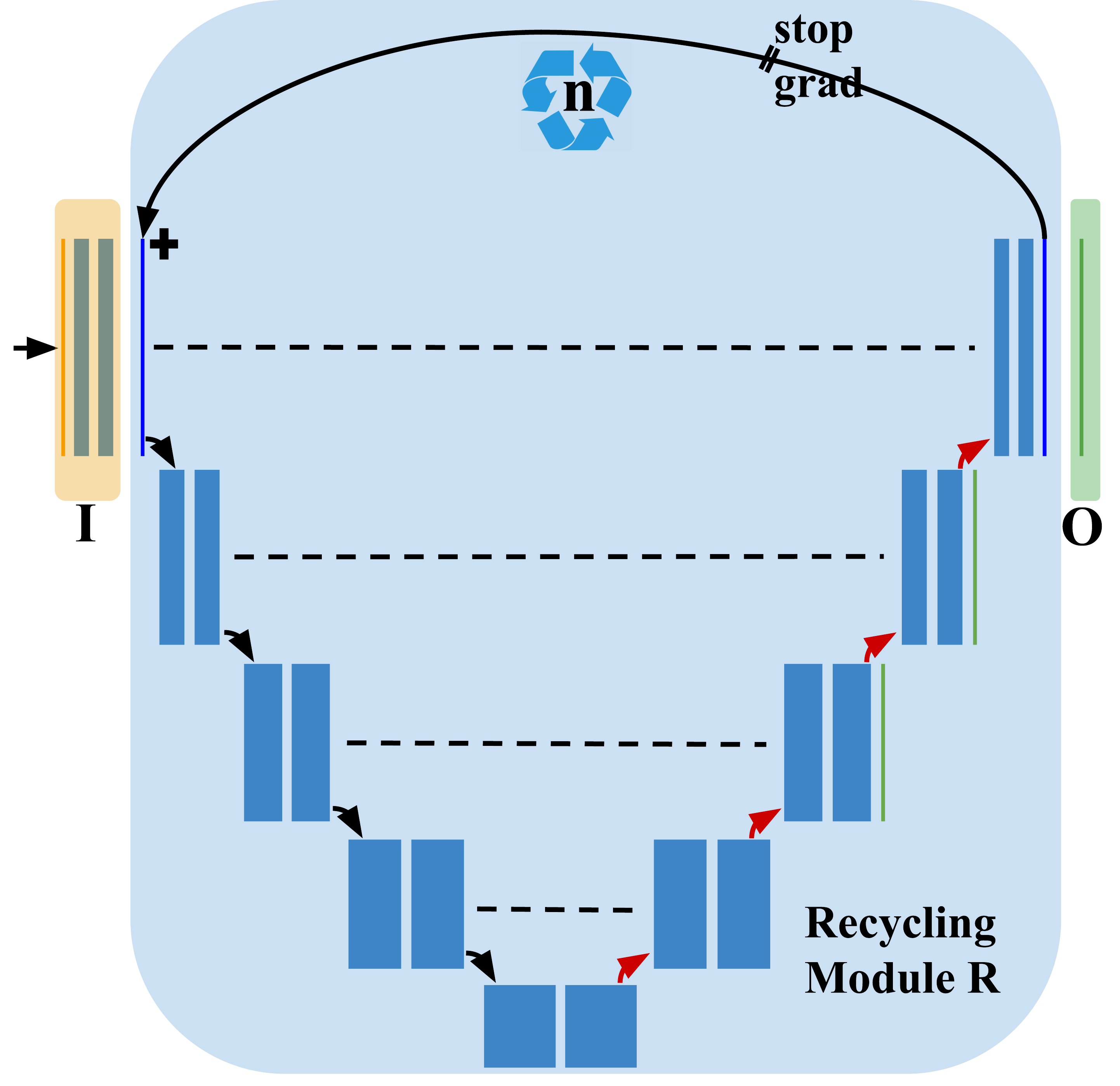}
\caption{Schematic overview of the proposed U-Net feature recycling. $n$ depicts the number of recycling cycles where the features close to the network's decoder are fed back into early encoder features. The letters I, R and O refer to the input projection, recycling module and output projection, as described in Algorithm \ref{alg:recycling}.}
\label{fig:recycling-scheme}
\end{figure}

Through our evaluation, we demonstrate that RecycleNet exhibits the remarkable property of refining its predictions iteratively, even beyond the iterations witnessed during training. The results clearly outperform the state-of-the-art segmentation methods across a range of segmentation benchmarks, demonstrating the promise of our approach.
At the heart of RecycleNet lies a trade-off - the opportunity to exchange increased computational time for a significant improvement in performance. For safety-critical applications, where the stakes are high, and the margin for error is virtually non-existent, this trade-off could be an important step in enhancing the reliability and precision of decision-making processes in neural networks. 


\section{Related Work}
\label{sec:related}
Neural networks' inability to refine initial predictions has been addressed in approaches which introduce varying degrees of additional complexity.

\paragraph{Refinement modules:} One straight-forward way to refine initial network outputs makes use of additional modules. In the context of image segmentation, these refinement modules typically act on segmentation maps \cite{recurrent-mask-refinement} or features close to the segmentation layers \cite{retinal-vessel-refinement} and make use of additional layers to refine a main network's outputs. This naturally introduces complexity by introducing additional parameters to the original network. In contrast, our suggested technique of latent feature recycling operates without requiring any additional parameters, thereby ensuring a more seamless integration in situations where the cost of extra parameters is prohibitive.

\paragraph{Recurrent Learning:} Another natural approach to refining initial network outputs is to cast refinement as a recurrent learning problem, with refinement steps as the temporal axis. To render typical computer vision network architectures as Recurrent Neural Networks, either the whole network \cite{feedback-networks} or key parts of the network \cite{recurrentUNet} are adapted. Alternatively, the recurrent learning can also be done on just the segmentation outputs \cite{recurrentPancreas}. While closely related to the proposed latent feature recycling, these approaches interfere strongly with the network architecture and due to their recurrent network formulation, come with substantial memory costs during training, which quickly become prohibitive for example in the context of medical image segmentation, where large 3D receptive fields are required to capture all relevant context for the task.

\paragraph{Multi-stage approaches:} To alleviate the memory costs connected to Recurrent Neural Networks, multi-stage approaches can be employed to refine previous stage segmentations \cite{auto-context} or learn based on the error feedback coming from the previous iteration \cite{pose-estimation}. While this does not come at the cost of substantially increased memory demands during training, multi-stage approaches often require multiple complete training cycles. Comparatively, our proposed method requires less training time since the additional forward passes are performed on a single sample.

\paragraph{Additional loss terms:} The ability to refine predictions has also been explored by repeatedly applying a given stateful network architecture and learning a stopping criterion using an additional loss term together with the task loss \cite{pondernet}. This, however, requires a careful balancing between loss terms which is not present in the proposed method.

\paragraph{Output recycling:} Recently, Jumper et al. \cite{alphafold2} have proposed a technique that uses output structures in the context of protein structure prediction to refine initial guesses without additional modules. They re-use outputs from certain transformer blocks \cite{vaswani2017attention} of the architecture over multiple iterations to refine the predicted structures. While this is closely related to our proposed technique, we show that recycling not only a part of the network architecture, but a whole convolutional segmentation network leads to refined predictions. Additionally, recycling features instead of outputs allows integrating this mechanism in a flexible way without special requirements w.r.t. the network architecture. Crucially, we also introduce a robust training schedule for recycling and demonstrate its importance for both reliable results across datasets, as well as an emergent convergence property where segmentation performance increases monotonously when increasing the number of cycles during inference.

\section{RecycleNet}
\label{sec:method}

Our proposed method, referred to as RecycleNet, relies on increasing the number of forward passes through a large part of the network (referred to as cycles), both during training and inference. To instill the capability to refine initial decisions over a number of such cycles, features close to the output are fed back into early layers of the neural network via a simple addition operation. \Cref{fig:recycling-scheme} depicts a schematic of the proposed latent feature recycling process. A given neural network architecture (here the U-Net \cite{ronneberger-unet}), can be partitioned into three disjoint parts: The input projection \textbf{I}, the recycling module \textbf{R} and the output projection \textbf{O} (see \cref{fig:recycling-scheme}). The recycling module \textbf{R} is not an additional module, but rather refers to the part of the architecture where features should be recycled. The recycling process, where recycled features are summed onto earlier feature representations, is repeated $N_c$ times during training, where $N_c$ is sampled uniformly from a predefined range (see \Cref{sec:schedule}). After the recycling process, the output projection produces the final output. However, each individual cycle can also be projected to a meaningful prediction, allowing for introspection and ensembling. The recycling mechanism is described in Algorithm \ref{alg:recycling}. We note that while it is, in general, possible to use gradients accumulated for more than one cycle, we find this to be impractical due to the memory demands involved. So instead, we only use gradients for the last iteration.

\begin{algorithm}[!ht]
\caption{Latent Feature Recycling: Training}
\label{alg:recycling}
    \SetAlgoLined
    \KwIn{Maximum number of cycles $N_{max}$,
    model input $x$, input projection I, recycling module R, output projection O} 
    Project input into recycling feature space: $z = I(x)$ \\
    Sample number of cycles: $N_c = RandInt(1, N_{max})$ \\
    Initialize recycling features as zeros: $r = 0$ \\
    \For{all cycles $i \in [1,...,N_c]$}{
        \uIf{$i < N_c$}{
            r = r.detach() $\#$ gradients only for last cycle
            }
        r = R(z, r) $\#$ 1 cycle
        }
    Project to output: $\hat{y} = O(r)$\\
        \Return $loss(\hat{y}, y)$
\end{algorithm}

Reusing the recycling features $r$ can be achieved in various ways. We propose a simple addition of normalized recycling features to the input projection, similar to the standard practice of adding position encodings in the context of Language Models \cite{vaswani2017attention}:

\begin{equation}
R(z, r) = R(z+norm(r))
\label{eq:additive-recycling}
\end{equation}

This approach requires the feature dimensions after the input projection to match the dimensions after the recycling module. This property is typically fulfilled for attention-based transformer architectures, as well as U-Nets \cite{vaswani2017attention}. In architectures, where this is not the case, other conditioning mechanisms, e.g. using projection layers, can be applied.
We propose the integration of latent feature recycling in the context of the U-Net \cite{ronneberger-unet}, a network architecture which is ubiquitous in medical image segmentation.\\
As depicted schematically in Figure \ref{fig:recycling-scheme}, we propose U-Net feature recycling by reusing features close to the output of the network's decoder at earlier layers, e.g. after the encoder's first convolutional block. This enables the network to revisit features based on which previous predictions would be computed, thus instilling the capability to iteratively refine early decision hypotheses over a number of cycles. 

\subsection{Robust Training Schedule}
\label{sec:schedule}

Feature recycling introduces a single new hyperparameter to an existing neural network training, the number of cycles $N_c$. This hyperparameter determines how many shots the network gets to refine initial predictions during training. At the beginning of the training, the initial predictions might be unreliable to the extent that there is little value in refining them step by step. To combat this, we introduce a robust recycling schedule, where during an initial warm-up phase, only a single cycle (no recycling) is used, therefore defaulting to standard network training during this period. Over time, we incrementally increase the range of possible cycles to allow for more and more refinement steps. As the number of cycles during training is not deterministic, the network is incentivized to distill useful information for each next recycling step in the recycled features. This helps learning an iterative refinement, even when only using gradients from the last cycle.

\section{Experiments and Results}
\label{sec:experiments}
In the following, we present the experiments and results based on the proposed U-Net feature recycling in the context of challenging medical image segmentation datasets. To reliably compare the proposed method with a strong baseline, we compare to the widely used nnU-Net and implement the proposed method in the same well-tested data pipeline and training framework \cite{isensee2021nnu}.

\subsection{Datasets and Evaluation}

To demonstrate the general effectiveness of the proposed U-Net latent feature recycling for medical image segmentation, we test the proposed method on a range of established segmentation tasks covering various dataset sizes, segmentation targets and task difficulties. These datasets include the Kidney Tumor Segmentation (KiTS 2019) dataset \cite{KiTS}, the Liver Tumor Segmentation task of the Medical Segmentation Decathlon \cite{decathlon}, the Multi-Atlas Labeling Beyond the Cranial Vault (BTCV) challenge \cite{BTCV} and the large-scale Abdominal Multi-Organ Benchmark (AMOS) for versatile medical image segmentation \cite{AMOS}. To also test the proposed method in the context of MRI tasks, we additionally include two tasks of the Combined Healthy Abdominal Organ Segmentation (CHAOS) challenge \cite{CHAOS}.\\
To assess segmentation performance, we use the average foreground Dice coefficient (DSC). It measures the overlap between predicted segmentation and ground truth, popular in medical image segmentation for its capacity at handling imbalanced datasets.

\subsection{Baseline Models}

To ensure a fair comparison, we implement all models in the nnU-Net framework \cite{isensee2021nnu}, which is regarded as the state-of-the-art framework for medical image segmentation and serves as the basis for numerous successful segmentation challenge contributions \cite{extending-nnU}. Naturally, the default nnU-Net is also included as a baseline. We also test against the best-performing model proposed in Wang, Yu et al. \cite{recurrentUNet} which still has a manageable memory demand when applied in the medical image segmentation domain, referred to as DRU. As the DRU was not originally tested in the context of 3D segmentation, we base our implementation on the authors' code and implement the model within the nnU-Net framework. This allows to make fair comparisons w.r.t. performance, memory demands and training time requirements.

\subsection{RecycleNet}
\label{sec:recycling-performance}

\begin{table*}[!h]
\centering
\begin{tabular}{l|c|c||c|c||c|c|c||c|c||c}
\multirow{2}{*}{\textbf{Model}} & \multicolumn{2}{c||}{\textbf{KiTS}} & \multicolumn{2}{c||}{\textbf{BTCV}} & \multicolumn{3}{c||}{\textbf{CHAOS}} & \multicolumn{2}{c||}{\textbf{AMOS}} & \multicolumn{1}{c}{\textbf{Liver T.}} \\ \cline{2-11} 
 & \multicolumn{1}{c|}{\textbf{CV}} & \textbf{Test} & \multicolumn{1}{c|}{\textbf{CV}} & \textbf{Test}  & \multicolumn{1}{c|}{\textbf{CV(T5)}} & \textbf{T3} & \textbf{T5} & \multicolumn{1}{c|}{\textbf{CV}} & \textbf{Test} & \textbf{CV} \\ \hline
nnU-Net & 89.29 & 89.04 & 82.96 & 87.21 & 94.77 & \textbf{93.49} & 91.47 & 88.58 & 90.68 & 78,84 \\
DRU \cite{recurrentUNet} & 89.58 & & 82.99 & & 91.62 & & & 88.62 & & \textbf{80.36} \\
RecycleNet & \textbf{90.26} & \textbf{89.11} & \textbf{83.75} & \textbf{87.80} & \textbf{94.92} & 93.48 & \textbf{91.85} & \textbf{88.77} & \textbf{90.82} & 79.88 \\
\end{tabular}
\vspace*{5mm}
\caption{Average DSC scores for 5-fold cross-validation (CV) and public leaderboard held-out test sets (Test). We compare the vanilla nnU-Net \cite{isensee2021nnu} with the proposed method and don't use post-processing on either model's predictions for a fair comparison. T3 and T5 refer to the test sets of two MRI tasks of the CHAOS challenge, selected to test the proposed method on a different modality than CT.}
\label{tab:5-fold}
\end{table*}

We implement the proposed RecycleNet as an adaptation of nnU-Net, making use of the same, well-tested, 3D full resolution U-Net architectures. For each dataset, we adopt the same feature recycling strategy, recycling the feature maps before the last convolutional layer and feed them back into the encoder after the first convolutional stage, see \Cref{sec:method}. Through this, we can keep the automatic configuration property offered by nnU-Net and roll out the same proposed recycling method for any given segmentation task. For a fair comparison between the proposed method and the baselines, we use the preprocessing pipeline of nnU-Net \cite{isensee2021nnu} for all experiments and evaluate on public leaderboards when possible. Where not possible, we employ an identical 5-fold cross validation. We evaluate single full resolution 3D models without ensembling or postprocessing. As the recycling training schedule, we make use of the schedule proposed in \Cref{sec:schedule}, starting with 1 cycle for the first 200 epochs, then gradually increasing the range from which the number of cycles is sampled by 1 every 200 epochs, to a maximum of 3. During inference, we increase the number of cycles to a maximum of 7 cycles to benefit from the property discussed in \Cref{sec:convergence} and always report metrics using the maximum number of cycles. 

Table \ref{tab:5-fold} shows the results comparing RecycleNet to the nnU-Net and DRU baselines on 5-fold cross validation as well as public leaderboard held-out test sets, where possible. We show clear performance improvements compared to both baselines across all evaluation datasets but the liver tumor segmentation dataset, where the DRU \cite{recurrentUNet} shows the best cross validation score. We note that although the differences between the individual methods seem small, they can be regarded as substantial improvements in the saturated performance domain that is medical image segmentation. We use a fixed recycling schedule and a fixed number of cycles during inference (determined based on the BTCV cross-validation, employed on all datasets), leaving all nnU-Net training hyperparameters untouched. We suspect that further improvements are possible by fine-tuning the training and recycling schedule on a given target dataset. From these results, we conclude that the proposed latent feature recycling represents an effective way to instill iterative refinement capabilities for even strong image segmentation models. The cost for this performance surplus is an increased training memory demand, as well as increased training and inference time (see \Cref{sec:memory}).

\subsection{Prediction Convergence of RecycleNet}
\label{sec:convergence}

\begin{figure*}[!ht]
\centering
\includegraphics[width=0.74\textwidth]{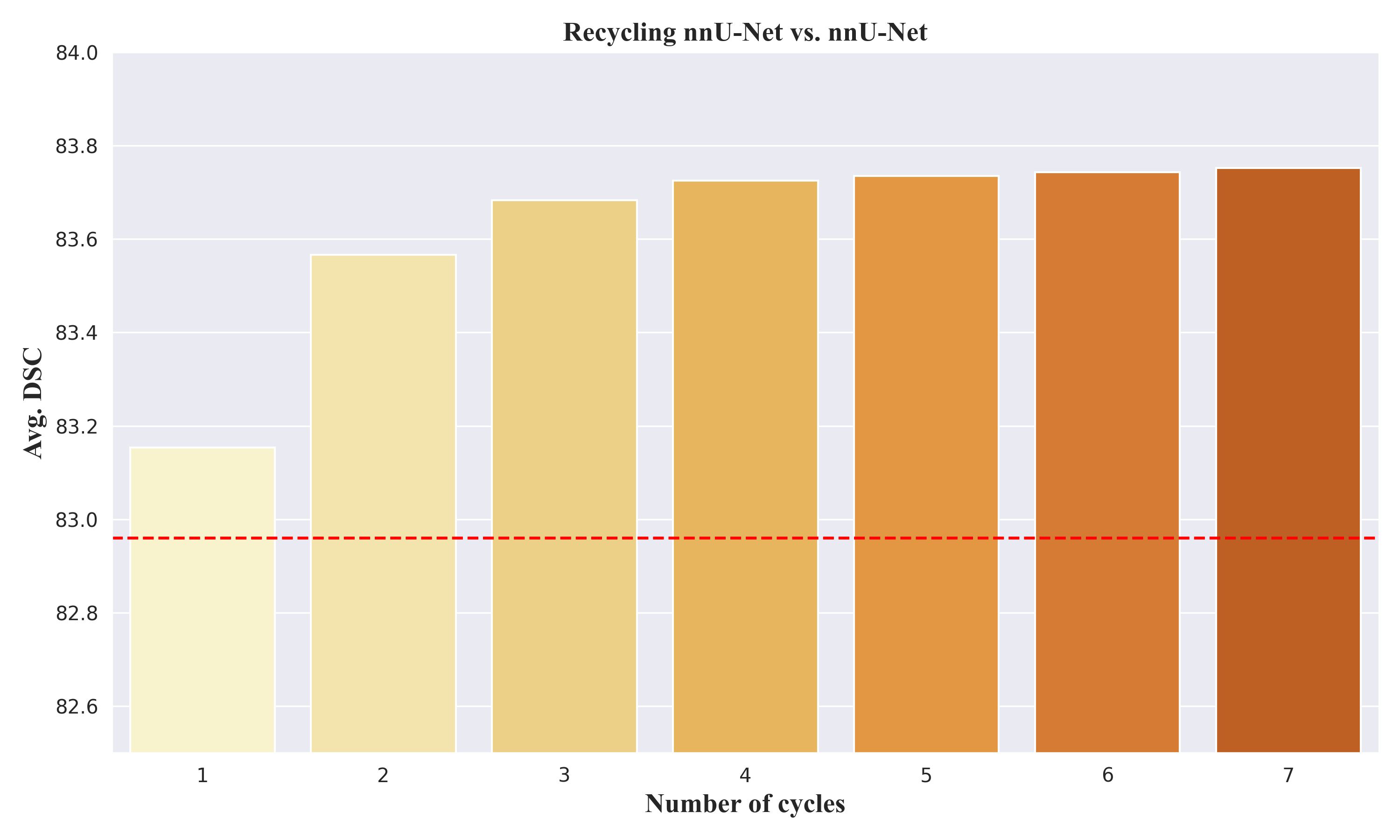}
\caption{Evaluation of the 5-fold cross validation performance of the proposed method on the BTCV dataset over number cycles in inference. The dotted red line represents the default nnU-Net performance, the yellow to orange bars represent the Recycling nnU-Net's performance for different numbers of cycles during inference.} 
\label{fig:recycling-convergence-dsc}
\end{figure*}

To test the iterative refinement capabilities of the proposed method, we investigate predictions over the number of cycles, both quantitatively and qualitatively. Figure \ref{fig:recycling-convergence-dsc} shows the performance in terms of average Dice score in a 5-fold cross validation on the BTCV dataset. 

\begin{figure*}[!ht]
\centering
\includegraphics[width=0.7\textwidth]{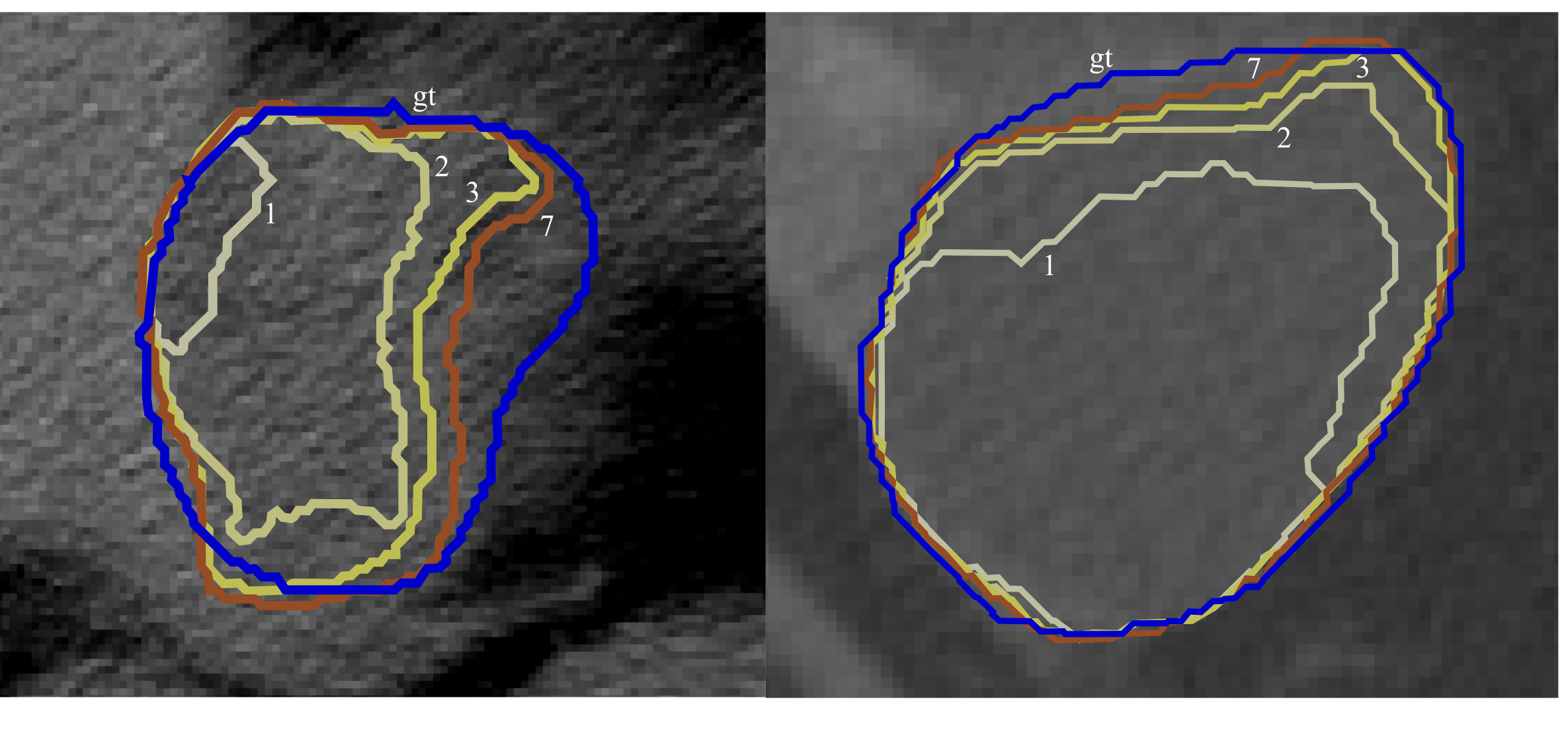}
\caption{Qualitative showcase of refined segmentations for liver tumor label in the Liver Tumor Segmentation Task \cite{decathlon} (left) and the gallbladder label in the BTCV dataset \cite{BTCV} (right). The initial prediction (light yellow) is iteratively refined (shades of yellow), converging towards an improved segmentation (brown). The predictions move closer to the ground truth label (blue) when increasing the number of cycles. Intermediate cycles omitted for visual clarity.} 
\label{fig:recycling-convergence-zoom}
\end{figure*}

We observe a monotonous performance increase when increasing the number of recycling cycles during inference, asymptotically converging towards a saturation DSC value. This suggests an interesting property arising from the introduction of feature recycling. We refer to this property as prediction convergence, where the model naturally converges towards a refined prediction. Surprisingly, even a single cycle can yield improvements compared with the baseline, as shown in Figure \ref{fig:recycling-convergence-dsc}. However, this observation is not consistent across datasets. We also note that the network learns to refine predictions even beyond the number of cycles seen during training, while still improving on prior predictions. This property is also reinforced when qualitatively inspecting the segmentation predictions the proposed method creates for single samples, as shown for samples of two segmentation tasks in \Cref{fig:recycling-convergence-zoom}. We see an increased prediction quality beyond the 3 cycles seen during training for both samples.

\subsection{Memory and Run Time}
\label{sec:memory}

\begin{figure*}[!h]
\centering
\includegraphics[width=0.75\textwidth]{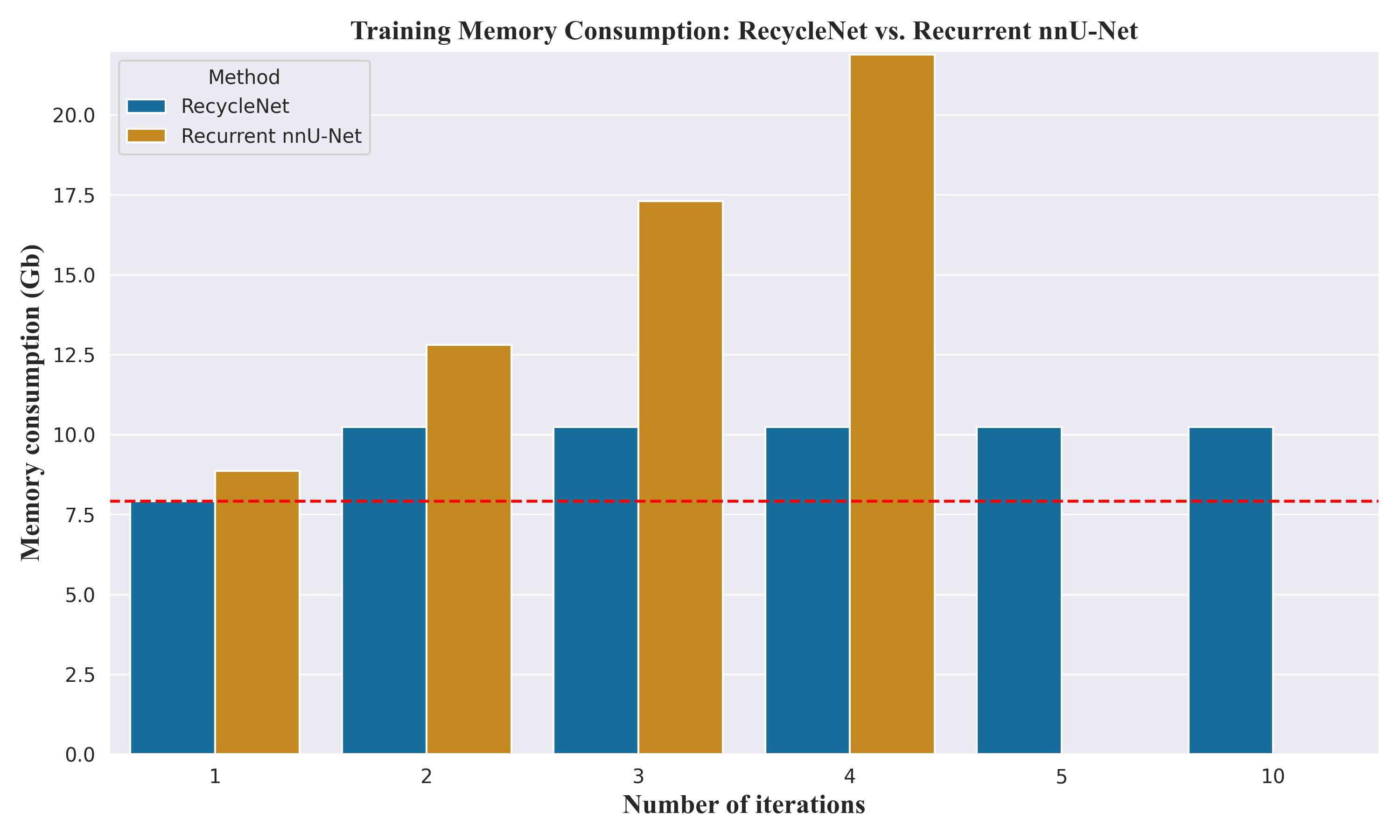}
\caption{Memory consumption during training on the BTCV dataset. We show the memory consumption for the proposed RecycleNet, as well as the Recurrent U-Net implemented in the nnU-Net framework. The memory consumption during training is displayed w.r.t. the number of iterations (recycling cycles or recurrent timesteps). While the proposed RecycleNet (blue) does come at the cost of a memory surplus, this surplus stays constant irrespective of the number of recycling cycles. Due to the recurrent nature of the Recurrent U-Net \cite{recurrentUNet} (light brown), the memory costs quickly become prohibitively large when increasing the number of recurrent timesteps during training. } 
\label{fig:recycling-memory}
\end{figure*}

In this section, we analyze the memory and training epoch time surplus of RecycleNet and the Recurrent U-Net baseline \cite{recurrentUNet}. Since we implemented both in the nnU-Net framework \cite{isensee2021nnu}, a fair comparison with an identical base network architecture, epoch definition, and underlying hardware can be ensured.

\paragraph{Training memory:}  \Cref{fig:recycling-memory} shows the memory consumption during training on the BTCV dataset w.r.t. the number of iterations. We notice a steep increase in memory consumption when training a Recurrent U-Net on medical image segmentation tasks. This increase stems from two sources: the additional recurrent module employed in the bottleneck of the network architecture and the recurrent training with backpropagation through time. In the context of 3D medical image segmentation, where the network architectures rely on large 3D inputs which lead to a large number of feature activations stored in memory for backpropagation, the latter contributes significantly to the total memory consumption. This leads to more than doubled memory requirements compared to a standard nnU-Net on this task. From this, we conclude that a recurrent U-Net does not efficiently scale to a larger number of recurrent time steps during training.\\
The proposed RecycleNet also shows a memory surplus during training, but due to the formulation described in \Cref{alg:recycling}, the memory demands saturate and don't grow with recycling cycles larger than two. In the experiments on the BTCV dataset, we see an increase in memory consumption of roughly 30\%. This surplus varies depending on the dataset at hand. We notice that future implementations should in principle be able to further reduce this memory surplus.

\paragraph{Training epoch time:} Considering the average training epoch times, we again report numbers on the BTCV dataset as a benchmark. Due to the 3D nature of medical image segmentation tasks, we note that data loading and augmentation can often become bottlenecks during training. However, the nnU-Net framework, along with current hardware enables alleviating this bottleneck, thus allowing for an unbiased measure of realistic epoch time changes stemming from the compared methods. While a typical nnU-Net epoch (defined as 250 mini-batch updates with 2 samples each) on the BTCV dataset takes 74 seconds, the Recurrent U-Net trained using 2 recurrent timesteps requires 165 seconds on average for each epoch. This is more than a 100\% increase in training time.

The proposed RecycleNet, however, comes with a much lower training time increase. While the later stage of the proposed recycling schedule leads to an epoch time increase of roughly 30\% during training (up to 96 seconds compared to nnU-Net's 74 seconds), this epoch time increase is non-existent during the first 200 epochs, where the recycling schedule only uses 1 cycle. Between epochs 200 and 400 of the proposed schedule, we measure an epoch time increase of roughly 15\%. When training for 1000 epochs, this leads to a total training time increase of about 20\% with an average epoch time of 89.4 seconds. 

In terms of memory and training time surplus, we see a non-negligible increase for the proposed RecycleNet. This increase, however, is much lower compared to the Recurrent U-Net baseline, rendering the RecycleNet much more accessible in the context of 3D medical image segmentation. We note that the inference time is also increased, since for each sample, $N_c$ forward passes have to be computed. However, in the context of time-consuming preprocessing and resampling, these additional forward passes only result in a minor inference time increase (roughly by a factor of 2 when increasing the number of cycles by a factor of 7).

\subsection{Ablation of Training Schedules}
\label{sec:ablation-schedules}

\begin{figure*}[!b]
\centering
\includegraphics[width=0.7\textwidth]{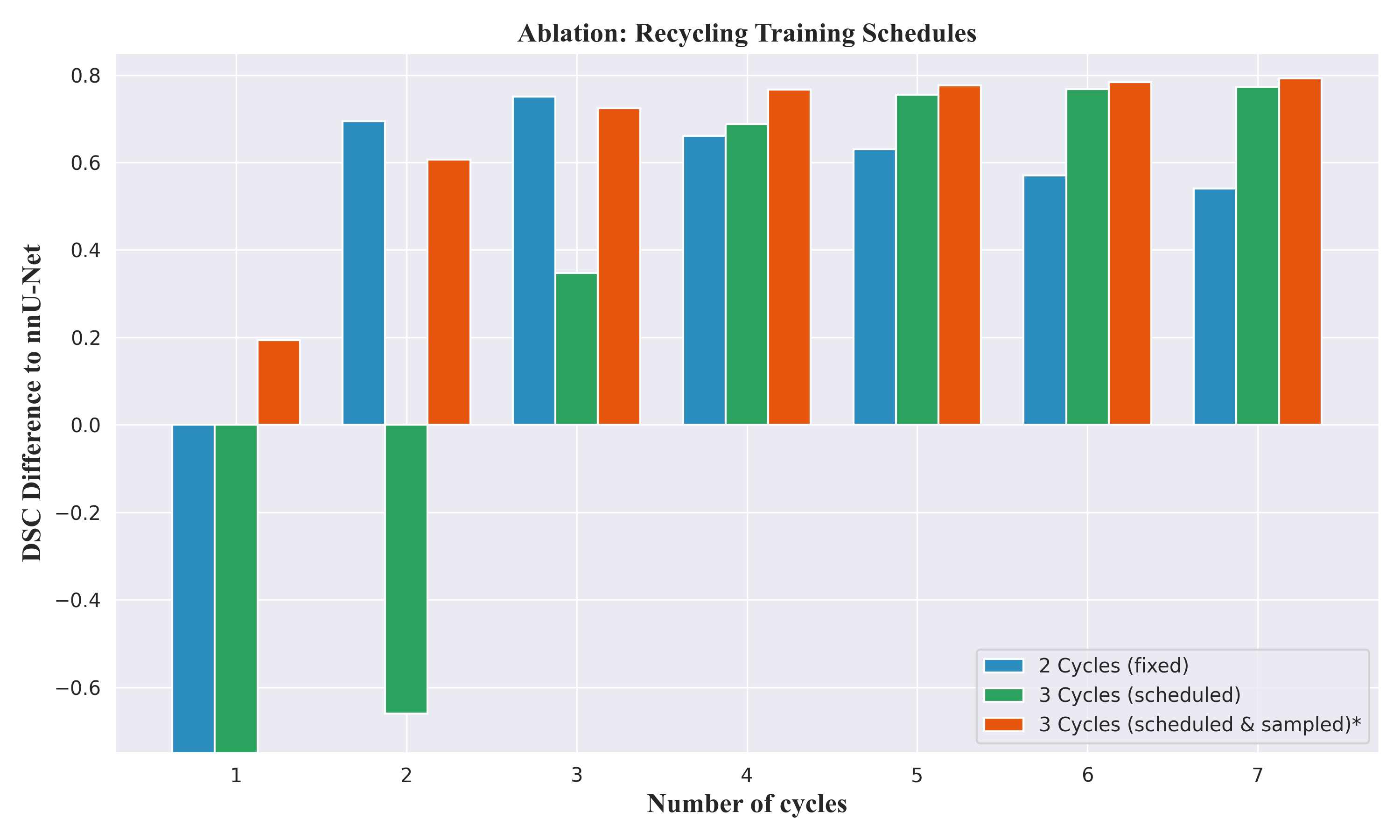}
\caption{Ablation of different recycling training schedules on the BTCV dataset. The difference in DSC (compared with nnU-Net) is shown w.r.t. the number of cycles used during inference. A static training schedule of 2 cycles is shown in blue, while an increasing training schedule (without sampling) is shown in green. The proposed schedule (see Section \ref{sec:schedule}), is marked and displayed in red.} 
\label{fig:recycling-ablation-schedules}
\end{figure*}

In this section, we investigate how different recycling training schedules affect the performance in the case of multi-organ segmentation on the BTCV dataset. \Cref{fig:recycling-ablation-schedules} shows differences in DSC score when compared with nnU-Net \cite{isensee2021nnu} in a 5-fold cross-validation. We show that while even using a fixed number of cycles during training (blue) can increase segmentation performance w.r.t. the baseline, this approach does not generalize well to a larger number of cycles during inference. Incrementally increasing the number of cycles over the course of the training process (green) does lead to a similar convergence property as discussed in \Cref{sec:convergence} when increasing the number of inference cycles beyond the maximum training cycles. However, due to the missing stochasticity in the number of cycles seen during training, the network forgets how to handle lower recycling numbers and performs poorly in those cases during inference. To ensure the observed convergence property described in \Cref{sec:convergence}, both the sampling and the incremental schedule components are important to achieve a reliable and strong prediction accuracy out of the box when confronted with unseen segmentation tasks. This quality is crucial for a wide applicability in the context of medical image segmentation.

\cleardoublepage
\section{Conclusion}

In this work, we propose a novel method for instilling iterative decision refinement capabilities into neural networks in a process we call latent feature recycling. This approach relies on minimal assumptions w.r.t. the network architecture and naturally leads to performance improvements, showing a convergence towards refined decisions. We demonstrate these capabilities using medical image segmentation as a showcase. In this context, latent feature recycling can improve on even strong models with a simple and robust recycling schedule. We observe that this schedule leads to iterative refinement capabilities, allowing to trade inference time for improved performance. These refinement capabilities even extend to a larger number of cycles than seen during training. We leave it for future work to explore the limits of this phenomenon for applications where even the smallest improvements in performance are worthwhile. As this approach is not limited to U-Nets or medical image segmentation, we expect further work to leverage latent feature recycling on a variety of neural network architectures and tasks. We note, however, that the proposed method comes with an additional computational cost through increased numbers of forward passes during training and inference. Compared with other iterative refinement approaches, we observe a much lower memory and training time surplus, making the RecycleNet a promising candidate e.g. in safety-critical applications where the additional memory and time requirements are not the limiting factor. However, this increased inference time potentially limits the application in time-critical applications. The proposed method also bears a resemblance to latent diffusion models \cite{latent-diffusion}, in that an iterative refinement process takes place in latent space, leading to a convergent refinement property during inference. Contrary to latent diffusion models, latent feature recycling covers multiple refinement steps also during training, whereas latent diffusion models don't allow multistep feedback during training. We leave it for future work to explore whether such multistep feedback is also beneficial in the denoising diffusion process.\\
\newpage
\paragraph{Acknowledgment:} This work was in part supported by the Helmholtz Association under the joint research school HIDSS4Health - Helmholtz Information and Data Science School for Health.\\
Part of this work was funded by Helmholtz Imaging, a platform of the Helmholtz Incubator on Information and Data Science.

\cleardoublepage

{\small
\bibliographystyle{ieee_fullname}
\bibliography{egbib}
}

\end{document}